\documentclass{article}

\usepackage[square,numbers]{natbib}

     \usepackage[preprint]{neurips_2019}

\usepackage[utf8]{inputenc} 
\usepackage[T1]{fontenc}    
\usepackage{hyperref}       
\usepackage{url}            
\usepackage{booktabs}       
\usepackage{amsfonts}       
\usepackage{nicefrac}       
\usepackage{microtype}      
\usepackage[toc]{appendix}

\usepackage[pdftex]{graphicx}
\DeclareGraphicsExtensions{.pdf,.jpeg,.png}
\graphicspath{{.}}

\usepackage{algorithm}
\usepackage{algpseudocode}

\title{Memory-Driven Mixed Low Precision Quantization For Enabling Deep Network Inference On Microcontrollers}

\author{%
  Manuele Rusci \\
  DEI, Universita' di Bologna\\
  \texttt{manuele.rusci@unibo.it} \\
  \And
  Alessandro Capotondi \\
  DEI, Universita' di Bologna\\
  \texttt{alessandro.capotondi@unibo.it} \\
  \And
  Luca Benini \\
  IIS, ETH Zurich \\
  DEI, Universita' di Bologna\\
  \texttt{lbenini@iis.ee.ethz.ch} \\
}

\begin{document}

\maketitle

\begin{abstract}
This paper presents a novel end-to-end methodology for enabling the deployment of low-error deep networks on microcontrollers. To fit the memory and computational limitations of resource-constrained edge-devices,  we exploit mixed low-bitwidth compression, featuring 8, 4 or 2-bit uniform quantization, and we model the inference graph with integer-only operations. 
Our approach aims at determining the minimum bit precision of every activation and weight tensor given the memory constraints of a device. This is achieved through a rule-based iterative procedure, which cuts the number of bits of the most memory-demanding layers, aiming at meeting the memory constraints. After a quantization-aware retraining step, the fake-quantized graph is converted into an inference integer-only model by inserting the Integer Channel-Normalization (ICN) layers, which introduce a negligible loss as demonstrated on INT4 MobilenetV1 models.
We report the latency-accuracy evaluation of mixed-precision MobilenetV1 family networks on a STM32H7 microcontroller. Our experimental results demonstrate an end-to-end deployment of an integer-only Mobilenet network with Top1 accuracy of 68\% on a device with only 2MB of FLASH memory and 512kB of RAM, improving by 8\% the Top1 accuracy with respect to previously published 8 bit implementations for microcontrollers.

\end{abstract}

\section{Introduction}
\label{Sec:intro}
Enabling machine learning on extreme-edge-devices is challenging due to their tight memory and computing power constraints. When envisioning smart sensors operating on batteries, the target power envelope must be below tens of mWs to guarantee a battery lifetime of years. This requirement impacts the system architecture design: adding computational units (e.g. floating-point units) or memory banks contributes increasing the complexity and the power cost, and hence the energy, of a system. 

Nowadays, microcontroller units (MCUs), such STMicroelectronics STM32 devices, feature an energy consumption compliant with the requirement of smart autonomous sensors and include energy-efficient computational units for running machine learning workloads. However, the typical size of the embedded memory cuts is limited to a few MB (a STM32H7 MCU features 2MB of FLASH memory) and the computation core (commonly a single ARM Cortex-M CPU) runs up to few hundreds of MHz. To boost the performance of this class of MCUs while leveraging the high flexibility of software-programmability, ARM recently released a software library, CMSIS-NN \cite{lai2018cmsis}, which enabled the efficient computation of deep networks on tiny microcontrollers. The optimized routines composing the library realize convolutional operations in fixed-point representations, to exploit instruction-level parallelism. Unfortunately, due to memory constraints, only a small set of relatively complex networks has been ported to the microcontroller domain yet \cite{zhang2017hello}. For what concerns models tailored for hard problems, e.g. image classification among the 1000 classes of Imagenet dataset, fitting them on MCU memory resources is still an open problem.

\begin{figure}[t]
  \centering
  \includegraphics[width=0.8\columnwidth]{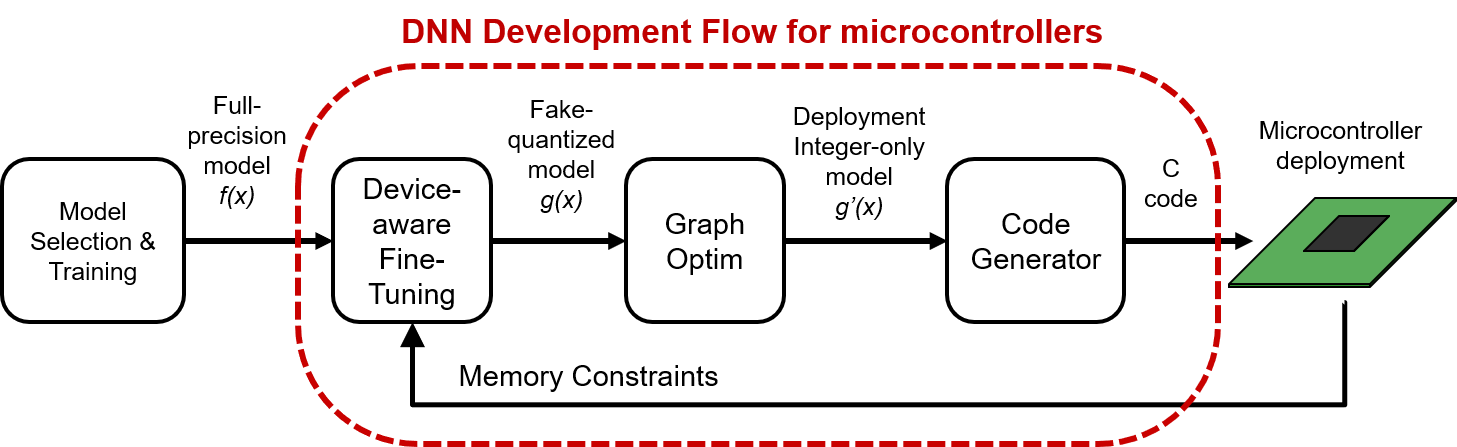}
  \caption{Design flow to bring deep neural networks into tiny microntrollers.}
  \label{fig:nnflow_micro}
  \vspace{-0.7cm}
\end{figure}

To address this problem, a crucial contribution comes from the recent work aiming at designing novel network topologies optimized not only in terms of accuracy but also for computational and memory costs \cite{howard2017mobilenets,ma2018shufflenet,wu2018fbnet}. In addition, a variety of compression techniques can be applied to further shrink a trained model. Among these, the quantization of either activations values and parameters to a low-bitwidth format, i.e. 8 bit or less,
is extremely effective because, besides reducing the memory footprint, it allows to operate with low precision integer operations, which can be efficiently mapped on the limited instruction-set of tiny microcontrollers. 
Figure \ref{fig:nnflow_micro} highlights the typical development flow to deploy a deep network design into a resource-constrained device. 
A pretrained network $f(x)$ is quantized by means of an initial device-aware fine tuning process, which can include also a re-training step.
The resultant fake-quantized model $g(x)$, emulating quantized values during the forward pass, is turned into an integer-only deployment model $g'(x)$ by means of an additional optimization step. Ideally, $\mathrm{loss}(g'(x)) = \mathrm{loss}(g(x)) = \mathrm{loss}(f(x)) $. 
The state-of-the-art methodology for training a quantized integer-only model is currently integrated within the Tensorflow framework, 
which shows a low accuracy degradation when targeting 8 bit implementations~\cite{jacob2018quantization}. 
This compression level is however not sufficient to bring complex models with high accuracy into memory-constrained microcontroller. As an example, a 8 bit MobilenetV1~\cite{howard2017mobilenets} with the highest accuracy requires more than 4 MB of embedded memory, which is prohibitive for the majority of microcontroller devices available.
To this end, a more aggressive sub-byte quantization methodology is needed, combined with novel techniques for deriving integer-only inference models.

In this work, we present a methodology for quantizing deep networks based on a mixed-precision scheme. The selection of the bit precision of every individual tensor is automated such as to satisfy the memory limitations of a given device.
Moreover, we improve the methodology~\cite{jacob2018quantization} for integer-only inference networks to support sub-byte per-channel quantization. Our experimental evaluation is conducted over the MobilenetV1 family networks on Imagenet~\cite{howard2017mobilenets}. We argue that this is a representative problem for tiny microntrollers, not yet solved~\cite{jain2019trained} and much harder than quantizing over-parameterized networks~\cite{choi2018pact}.

This paper places the following contributions: i) We introduce the Integer Channel-Normalization (ICN) activation layer to achieve an efficient conversion of the fake-quantized graph, also exploiting per-channel quantization and optimized quantization-aware training strategies, into an integer-only deployment graph. ii) We present a mixed-precision quantization methodology driven by the memory constraints of a target architecture, which aims at selecting the bit precision of every weight and activation tensor of an integer-only network. iii) We studied the latency-accuracy tradeoff on iso-memory mixed-precision networks belonging to the MobilenetV1 family when running on a STM32H7 microcontroller device.

Our methodology demonstrates an integer-only deployment of a MobilenetV1 network on a STM32H7 microcontroller with 68\% Top1 accuracy, which is 8\% higher than previous reported 8 bit integer-only implementations~\cite{jacob2018quantization}.

\section{Related Work}
\label{Sec:rel_work}
\textbf{Quantized Neural Networks.}
Early works on quantization of deep networks targeted 16 bits fixed-point implementations~\cite{lin2016fixed}, which result in an almost lossless approximation of full-precision trained networks, or extreme binarized networks, which, despite the fascinating low-computational and memory requirements, showed major accuracy losses when applied on image classification benchmarks~\cite{courbariaux2016binarized,rastegari2016xnor}. 
Several studies demonstrated that 8 bit quantization of weights and activations results in a good trade-off between latency, compression and a near-zero accuracy degradation, also if applied to efficient Imagenet classification networks~\cite{jacob2018quantization,migacz20178,jain2019trained}. Among the employed methodologies, TensorRT~\cite{migacz20178} approximates the parameters tensor by the minimization of the KL divergence metric between quantized and full-precision values. On the contrary,~\cite{jacob2018quantization} quantizes values within a range defined by the tensor min and max values. Concerning activations, the PACT approach~\cite{choi2018pact} demonstrated the highest efficiency by leveraging backpropagation to learn the quantization ranges. 
Recently, to fit stringent memory requirements, more aggressive sub-byte precision quantization approaches, i.e. less than 8 bit, are under investigation~\cite{choukroun2019low,jain2019trained,esser2019learned,krishnamoorthi2018quantizing,liu2019learning}. 
The works~\cite{jain2019trained, esser2019learned} exploits learning-based approaches for determining the quantization ranges of activation and weights at low-bitwidth precision.
State-of-the-art accuracy level on the efficient MobilenetV1 model has been reported by~\cite{krishnamoorthi2018quantizing,liu2019learning}, by making use of per-channel quantization when moving to 4 bits precision. 
It is also worth to mention as non-uniform quantizers have resulted as the best approximators when reducing the bit precision~\cite{zhang2018lq,wang2018haq,han2015deep}. However, a high-precision (floating point) arithmetic is needed on uncompressed values within the datapath, hence these methods results not suitable for the microcontroller domain.
In this work, we leverage existing techniques and show the insights, concerning either computational and memory aspects, when bringing fake-quantized networks to the integer-only arithmetic domain, which is not taken into consideration by this class of works.

\textbf{Mixed Low Precision Quantization.} 
Mixed-precision techniques make use of multiple bit precision throughout a quantized network, motivated by the fact that a lossy and aggressive linear cut is not necessary to reach a given compression rate.
The method~\cite{fromm2018heterogeneous} targeted per-pixel binarization based on a defined tensor mask. 
Despite achieving an extreme quantization level, a per-pixel quantization cannot be efficiently handled on a microcontroller, due to the control-based nature of the required dataflow.
The HAWQ~\cite{dong2019hawq} method relies on a second order Hessian metric to define prioritization of tensor's bit precision to reduce, but without choosing the optimal per-tensor quantization level. 
On the same direction, HAQ~\cite{wang2018haq} dynamically explores multiple low-bitwidth precision at training time by means of reinforcement learning. When optimizing for memory constraints, a non-uniform quantization is used. 
Compered to this, our methodology for bit precision selection applies statically, before quantization-aware retraining, and it is based on a rule-based iterative procedure. 
Both \cite{dong2019hawq} and \cite{wang2018haq} reports superior accuracy than ours when compressing networks to a 1MB of memory footprint, but they include non-uniform clustering quantization of floating-point parameters, therefore not fully-comparable with our work in terms of microcontroller readiness, as current MCUs are not equipped with the hardware needed for manipulation and computation on these data formats. 

\textbf{Deep networks for resource-constrained devices.} To bridge the gap between the complexity on deep networks and the limitations of resource-constrained devices, device-aware optimization strategies have also been presented. 
The work~\cite{blott2018finn} introduced FINN-R to quantize and deploy a generic model into constrained FPGA architectures. 
Their quantization approach makes use of integer thresholds ~\cite{umuroglu2017streamlined,gao2018ifq,rusci2018work} for data compression. This method enabled a lossless integer representation of a fake-quantized networks, but demands larger memory footprint with respect to our proposed method. 
In contrast, the integer-only deployment in~\cite{jacob2018quantization} presented a compact fixed-point 8 bit quantization strategy, which performs the folding of batch-normalization and scaling factors into weights before applying a uniform quantizer. Additionally, per-layer fixed-point parameters are needed for adapting the dynamic range when passing data from a layer to the next one. In contrast with this work, our methodology generalizes the deployment process when a more effective quantization strategy is used, i.e. per-channel mixed-precision quantization. 



\section{Background on Low-Bitwidth Quantization}
\label{Sec:qnn}


The quantization process aims at quantizing either the network parameters and the activations values, i.e. the temporary input and output values of the network layers. While the parameters can be quantized just before the inference (forward) pass~\cite{migacz20178}, the quantization of the activations requires the insertion of \textit{fake-quantized} activation layers within the network graph. These additional layers are responsible for recording the activation range statistics, optionally via backpropagation~\cite{choi2018pact}, and apply quantization during the forward pass depending on the collected statistics. Because of injected quantization noise, the original full-precision network $f$ is approximated with the correspondent fake-quantized function $g$. A quantization-aware retraining of a fake-quantized model is essential to recover accuracy, especially when low-bitwidth precision is employed~\cite{jacob2018quantization}.

In the remainder of the paper we only focus on uniform quantization because its arithmetic is naturally supported the instruction-set of general-purpose programmable MCUs.
Hence, without loosing generalities, any tensor $t\in \mathbb{R}^N$, either representing weights or activations or only a subset of them, can be quantized across the range $[a,b]$ with a given number of $Q$ bits~\cite{jacob2018quantization} as:
\begin{equation} \label{eq:quant}
    T = \mathrm{quant}(t) = \mathrm{round}( \frac{\mathrm{clamp}(t,a,b)}{S_t}) S_t \quad, \qquad
    S_t = \frac{b-a}{2^{Q}-1} 
\end{equation}
Equation (\ref{eq:quant}) derives from the mapping:
\begin{equation} \label{eq:quant2}
    t = S_t \cdot (T - Z_t)
\end{equation}
where $Z_t$ is a bias parameter required to shift the numeric domain of the quantized tensors $T$ into $[0,2^{Q}-1]$ or $[-2^{Q-1},2^{Q-1}-1 ]$ ranges, representative of the UINT-Q and INT-Q datatypes.
If $Z_t$ is constrained to zero, e.g. when $a = -b, b>0$, the quantization range is symmetric. 

In the case of weights, the parameters $a$ and $b$ can be computed as the min and max values of a tensor~\cite{jacob2018quantization} or by means of more sophisticated statistic analysis~\cite{migacz20178} or via backpropagation~\cite{choi2018pact}. A Per-Layer (PL) quantization exploit single values $a$ and $b$ for the whole full-precision tensor, hence the Equation~\ref{eq:quant} is applied layer-wise. A Per-Channel (PC) procedure results more effective by independently approximating a given tensor along the outer dimension~\cite{krishnamoorthi2018quantizing}. This corresponds to compute the $a$ and $b$ parameters in correspondence of any output channel of the tensor. 

To determine the quantization range of the activation values, statistics can be collected at training time during the forward pass, or against a specific calibration dataset. The PACT strategy demonstrated the effectiveness of learning $b$ via backprogation while $a=0$ to reproduce the non-linearity of the ReLU function. In our implementation, the $\mathrm{round}(\dot)$ of Equation~\ref{eq:quant} is replaced by $\mathrm{floor}(\dot)$ because of the lighter software implementation (the operand gets simply truncated, i.e. a shift operation), becoming: $\mathrm{quant\_act}(x) = \mathrm{floor}( \frac{\mathrm{clamp}(x,0,b)}{S_x})
    ,S_x = \frac{b}{2^{Q}-1}$.

\section{Integer-Only Inference}
\label{Sec:deploy}
Previous work~\cite{jacob2018quantization} discussed the training and integer-only deployment of a fake-quantized network with 8 bit per-layer quantization. 
The weight quantization is applied after folding the batch-norm parameters into the convolutional weights.
However, when reducing the bit precision below 8 bit using per-layer quantization, the folding process itself can lead to accuracy drop because it can drastically affects the range of the parameters to quantize. As a reference, Table \ref{tab:int4_mobilenet} shows the collapse of the training process for INT4 MobilenetV1 with the folding of the batch-norm parameters enabled.

With the aim of an integer-only deployment, we extend~\cite{jacob2018quantization} to a) prevent the folding of batch normalization parameters into convolutional weights and b) support per-channel low-bitwidth weight quantization.  
We observe that any fake-quantized network's sub-graph composed by a convolutional layer, a batch-normalization layer and a fake-quantizer activation module can be modeled by the transfer function: 

\begin{equation} \label{eq:1}
    y = \mathrm{quant\_act} ( \frac{\phi - \mu }{\sigma} \cdot \gamma + \beta )
\end{equation}

where $\phi=\sum x \cdot w$ is the output of a full-precision convolution and $\mu,\sigma,\gamma, \beta$ are channel-wise full-precision parameters of a batch normalization layer. It is worth to note that this kind of formulation holds for any feature-wise or layer-wise scaling factor applied to the convolution's output tensor.

When applying a per-layer quantization of either input/output activations and weights, the Rule~\ref{eq:quant2} is injected into Equation~\ref{eq:1} that becomes:

\begin{equation} \label{eq:2}
    Y = \mathrm{quant\_act} (Z_y +  \frac{S_i S_w}{S_o} \frac{\gamma}{\sigma}(\Phi + \frac{1}{S_i S_w} \cdot (B - \mu + \beta \frac{\sigma}{\gamma} )) )
\end{equation}

where $\Phi=\sum (X-Z_x) \cdot (W-Z_w)$ is the integer output of a low-bitwidth convolution.
We define the arrays $B_q=\mathrm{round} (\frac{1}{S_i S_w} \cdot (B - \mu + \beta \frac{\sigma}{\gamma}) )$, i.e. the quantized bias,  and $M=\frac{S_i S_w}{S_o} \frac{\gamma}{\sigma}$. As done by~\cite{jacob2018quantization}, each element $m_i$ of $M$ can be decomposed as $m_i = m0_i \cdot 2^{n0_i} $, where $m0_i$ is a signed fractionary fixed-point number with $0.5 \le \mathrm{abs}(m0_i) < 1.0 $. For the sake of notation, we indicate as $M_0$ and $N_0$ the two vectors such as $M = M_0 \cdot 2 ^ {N_0}$.
Given this, Equation~\ref{eq:2} can be rewritten as:

\begin{equation} \label{eq:3}
    Y = \mathrm{clamp} (Z_y +  \mathrm{floor ( M_0 \cdot 2 ^ {N_0} \cdot (\Phi + B_q) ), 0, 2^{Q} -1 )} 
\end{equation}

Note that every value in Equation~\ref{eq:3} is an integer or a fixed-point value, so that a quantized convolutional layer can be computed with an integer-only arithmetic. Since the static parameters $M_0,N_0,B_q$ vary along the channel dimension, we name this activation function (Equation~\ref{eq:3}) as \textbf{Integer Channel-Normalization activation}, indicated as $\mathrm{ICN}$.
If weight parameters get quantized per-channel (PC), i.e. every output channel weight bank has its own $S_w$ and $Z_w$ values, Equation (\ref{eq:3}) still holds after deriving the $B_q$, $M_0$ and $N_0$ vector parameters accordingly.

\subsection{Memory Requirement} \label{sec:mem_req}
Table~\ref{tab:mem_qlayer} schematizes the memory requirements to compute the transfer Function~\ref{eq:3}, considering both per-layer (PL) or per-channel (PC) quantization and the ICN layer. The table reports the amount of parameters of a convolution operation with a $k_w \ \mathrm{x} \ k_h$ receptive field, $c_I$ input channels and $c_O$ output channels. The weight-parameters are stored in memory as UINT-Q, where Q denotes the number of bits, so that the represented numeric domain corresponds to [0, $2^Q-1$]. $Z_x$, $Z_w$ and $Z_y$ are in a UINT8 format ($Z_w$ as INT16 if PC is applied), $B_q$ and $M_0$ are stored as INT32 and $N_0$ is a INT8 array. For comparison purpose, Table~\ref{tab:mem_qlayer} reports also the higher memory requirement of a quantized convolutional layer if using the thresholding method proposed by~\cite{umuroglu2017streamlined,gao2018ifq}, which exponentially increases with $Q$. 

\begin{table} 
\footnotesize
  \caption{Memory Requirements of a Quantized Convolutional Layer}
  \label{tab:mem_qlayer}
  \centering
  \begin{tabular}{l |c  | c | c | c | c | c | c | c}
    \toprule
    Label  & $Z_x$ &  Weights & $Z_w$ &  $B_q$ &  $M_0$ &  $N_0$ &  $Z_y$ & Thr \\
    \midrule

    PL+FB \cite{jacob2018quantization} &  1 & $c_O \cdot k_w \cdot k_h\cdot c_I$ & 1 & $c_O$ & 1 & 1 & 1 & - \\
    
    PL+ICN ~(our) &  1 & $c_O \cdot k_w \cdot k_h\cdot c_I$ & 1 & $c_O$ & $c_O$ & $c_O$& 1  & -\\
    
    PC+ICN~(our)  &  1 & $c_O \cdot k_w \cdot k_h\cdot c_I$ & $c_O$ & $c_O$ & $c_O$& $c_O$& 1 & - \\
    \midrule
    PC+Thresholds~\cite{umuroglu2017streamlined,gao2018ifq} & 1 & $c_O \cdot k_w \cdot k_h\cdot c_I$ & $c_O$ & - & - & - & 1 & $c_O \cdot 2^{Q} $ \\  
    \bottomrule
  \end{tabular}
\vspace{-0.5cm}
\end{table}

\section{Memory-Driven Mixed Low Precision Methodology for MCU Deployment}
\label{sec:mem-driven}

To run deep networks on microcontrollers, the memory footprint is a stringent constraint. 
Given common microcontroller architectures~\cite{zhang2017hello}, we distinguish: 
\begin{itemize}
\item \textbf{Read-Only (RO) Memory}, to store frozen inference parameters, i.e. parameters that will not change during the lifetime of a smart device.
\item \textbf{Read-Write (RW) Memory}, to store temporary values, i.e. input and output of any quantized convolutional layer that depends on the current sensor data.
\end{itemize}
At any step of the inference pass, a pair of temporary activation tensors, i.e. the input and output of a layer, and the whole set of fixed parameters must be present in the memory. If considering a network of $L$ stacked quantized convolutional layers and a device with $\mathrm{M_{RO}}$ and $\mathrm{M_{RW}}$ memory budget (expressed in bytes), the above requirement is translated as:

\begin{equation} \label{eq:wmem}
    \sum_{i=0}^{L}\mathrm{mem}(w_i,Q_w^i)+M_{T_A^i} \le \mathrm{M_{RO} }
\end{equation}

where $i$ indicates the i-th quantized convolutional layer and $\mathrm{mem(t, Q)}$ returns the memory footprint of a tensor $t$ with bit precision $Q$. $M_{T_A^i}$ is the memory footprint of the additional set of layer's static parameters (see Table \ref{tab:mem_qlayer}) with datatype detailed in Section \ref{sec:mem_req}. Concerning activation values:

\begin{equation} \label{eq:amem}
    \mathrm{mem}(x_i,Q_x^i) +\mathrm{mem}(y_i,Q_y^i) \le \mathrm{M_{RW}} \qquad i=0,..,L-1
\end{equation}

Our methodology aims determining the bit precision $Q_w^i,Q_x^i,Q_y^i$   of any input $x_i$, output $y_i$ and weight $w_i$ tensor of the $i$-th layer, to match the memory constraints (\ref{eq:wmem}) and (\ref{eq:amem}). 
Only the values of $Q=\{2,4,8\}$ are admittable solutions; $Q_x^0$ is fixed to 8. 
Note that $y^i \equiv x^{i+1}$, hence fixing $Q_y^i$ is equivalent to set $Q_x^{i+1}$.
Initially, the bit precision of every tensor is set as $Q=8$. Algorithm \ref{alg:activ} and Algorithm \ref{alg:weight} reports the pseudo-code of the procedure to cut the bit precision of, respectively, activations and weights, under the hypothesis that exists a solution that satisfy (\ref{eq:wmem}) and (\ref{eq:amem}). 
The procedure in Algorithm \ref{alg:activ} iterates over the $L$ quantized convolutiona layers in a forward and backward fashion: the bit precision of output tensors $Q^i_y\equiv Q^{i+1}_x$ are cut during the forward pass, reductions of the input tensors' precision $Q^i_x\equiv Q^{i-1}_y$  are applied during the backward pass. Any cut consists of reducing the bit precision by a single step, i.e. from 8 to 4 and from 4 to 2 bits, and it is applied if the number of bits of the intended tensor (output during forward or input during backward) is lower or equal, but with a higher footprint, than the other activation tensor of the i-th layer.

\begin{algorithm}[t]
\footnotesize
\caption{Cut Activation Bits}
\label{alg:activ}
\begin{algorithmic}[1]
    \Require a fake-quantized network $g$ of $L$ stacked quantized convolutional layers, a $\mathrm{M_{RO}}$ memory constraint, a $Q_{a,min}$ minimum quantization level
    \Ensure the bit precion $Q_x^i,Q_y^i,i=0,..L-1$ to satisfy (\ref{eq:amem})
        \State $Q_y^i\equiv Q_x^{i+1} \gets 8 \qquad i=0,..L-1$  \Comment{Initialization}

    \While{(\ref{eq:amem}) is not True for every layer} \Comment{Stop Condition}
        \For{$i = 0$ to $L-2$} \Comment{Forward pass}
            \While{$\mathrm{mem}(x_i,Q_x^i  )+\mathrm{mem}(y_i,Q_y^i )  >\mathrm{M_{RW}}$ \textbf{AND} $\mathrm{CutBits}(x_i,Q_x^i,y_i,Q_y^i)$}
                \State $Q_y^i $ and  $Q_x^{i+1}$ are decremented by one step 
            \EndWhile

        \EndFor
        \For{$i = L-1$ to $1$}  \Comment{Backward pass}
            \While{$\mathrm{mem}(x_i,Q_x^i  )+\mathrm{mem}(y_i,Q_y^i ) >\mathrm{M_{RW}}$  \textbf{AND} $\mathrm{CutBits}(y_i,Q_y^i,x_i,Q_x^i)$}
            \State $Q_x^i $ and  $Q_y^{i-1}$ are decremented by one step
            \EndWhile
        \EndFor
    \EndWhile
    
  \State  
  \Function{CutBits}{$x_1,Q_{x_1},x_2,Q_{x_2}  $} \Comment{Return True if $Q_{x_2}$ have to be decremented }
  \If{ $Q_{x_2} > Q_{a,min}$}  
  
    \If{$ Q_{x_2} > Q_{x_1} $ \textbf{OR} ( $Q_{x_2}== Q_{x_1}$  \textbf{AND} $\mathrm{mem}(x_2, Q_{x_2}  ) > \mathrm{mem}(x_1, Q_{x_1} ) $ ) }
        \State \Return True
    \EndIf
  \EndIf
  \State \Return False
  \EndFunction
\end{algorithmic}
\end{algorithm}

Algorithm~\ref{alg:weight} details the iterative procedure for cutting bits of the weights parameters. At any iteration, a layer score $r_i$ is computed as the ratio between the layer's footprint of the i-th layer and the total occupation. Among the highest scores $r_i$ within a $\delta$ margin, the layer with the lowest layer's index is selected for the cut. This heuristic rule is intended to favorite the cut of central layers with respect to the last layers, usually more critical for what concern quantization.

\begin{algorithm}[t]
\footnotesize
\caption{Cut Weights Bits}
\label{alg:weight}
\begin{algorithmic}[1]
    \Require a fake-quantized network $g$ of $L$ stacked quantized convolutional layers, a $\mathrm{M_{RW}}$ memory constraint, a $Q_{w,min}$ minimum quantization level, a $\delta$ margin
    \Ensure The bit precision $Q_w^i,i=0,..L-1$ to satisfy (\ref{eq:wmem})
        \State $Q_w^i \gets 8$ \Comment{Initialization }
    \While{$\sum_{i=0}^{L-1}{\mathrm{mem}(w_i,Q_w^i) + M_{T_A^i} }>\mathrm{M_{RO}}$ }
        \State Compute $r_i = \mathrm{mem}(w_i,Q_w^i ) / \sum_{i=0}^{L-1}{\mathrm{mem}(w_i,Q_w^i)}$ for every layer with $Q_{w_i} >  Q_{w,min}$ 
        \State Find $R = \max_i r_i$
        \State Among the layers with $r_i > (R - \delta$), select the k-th with the smallest index $i$
        \State $Q_w^k $ is decremented by one step
    \EndWhile

\end{algorithmic}
\end{algorithm}

\section{Experimental Results}
\label{Sec:exp}

We run experiments on the MobilenetV1 family networks~\cite{howard2017mobilenets} on Imagenet using the PyTorch framework. In the following, a MobilenetV1 model is referred with a label $x\_y$, where $x=\{128,160,192,224\}$ is the spatial resolution of the input data and $y=\{0.25,0.5,0.75,1.0\}$ refers to the width channel multiplier. The quantization-aware retraining starts from pre-trained weights\footnote{Pretrained weights are downloaded from \url{https://github.com/tensorflow/models/blob/master/research/slim/nets/mobilenet_v1.md}}. Every training session executes on a compute node equipped with 4 NVIDIA-Tesla P100 GPUs for 8 hours. ADAM is chosen as optimizer with an initial learning rate of 1e-4, which is decreased in a fixed schedule to 5e-5 and 1e-5 at, respectively, the 5th and 8th epoch. Running statistics and learned parameters of batch-normalization layers are frozen after the first training epoch. Batch size is 128. An asymmetric uniform quantization is applied on weights: the PACT method is used in case of PL quantization while min/max statistics are employed in case of PC quantization. PPQ~\cite{liu2019learning} is applied for refining pre-trained weights before the quantization-aware retraining. Folding of batch-normalization parameters into weights, when applied layer-wise, starts from the 2nd training epoch. Activations are quantized with the PACT strategy.
%
%
\begin{table}[t]
\footnotesize
  \caption{Integer-Only MobilenetV1\_224\_1.0}
  \label{tab:int4_mobilenet}
  \centering
  \resizebox{0.6\columnwidth}{!}{%
  \begin{tabular}{ l | c  | c  }
    \toprule
    Quantization Method & Top1 Accuracy & Weight Memory Footprint \\
    \midrule
    Full-precision \cite{jacob2018quantization} & 70.9\% & 16.27 MB  \\
    PL+FB INT8 \cite{jacob2018quantization} & 70.1\% & 4.06 MB \\
    \midrule

    PL+FB INT4 (our) & 0.1\% & 2.05 MB  \\
    
    PL+ICN INT4 (our)  &  61.75\% & 2.10 MB \\
    
    PC+ICN INT4 (our)  & 66.41\% & 2.12 MB  \\
    
    \midrule

    PC W4A4 \cite{liu2019learning} & 64.3\% & -  \\
    PC W4A8 \cite{krishnamoorthi2018quantizing} & 65\% & -  \\
    \midrule

    PC+Thresholds INT4 (our) & 66.46\% &  2.35 MB \\
    \bottomrule
  \end{tabular}
  }\vspace{-0.5cm}
\end{table}

To proof the effectiveness of the ICN layers, we apply our quantization approach to a MobilenetV1 224\_1.0 model and we measure the accuracy achieved by a 4 bit integer-only implementation.
Table~\ref{tab:int4_mobilenet} reports the accuracies for the following strategies: PL+FB stands for per-layer quantization and folding of batch-norm parameters into weights, PL+ICN indicates per-layer quantization with ICN layers and PC+ICN refers to per-channel quantization with ICN layers.
First we can note that only thanks to the proposed ICN layers, the folding of the batch-norm parameters, which causes the collapse of the training process (PL+FB INT4), can be avoided, therefore enabling the convergence of the training algorithm (PL+ICN INT4 and PC+ICN INT4).
Secondly, the insertion of the ICN layer introduces an almost negligible accuracy drop of 0.3\% on PL+ICN and 0.05\% on PC-ICN with respect to the fake-quantized graph.
Moreover, by means of PC quantization, the accuracy of our 4 bit model is higher than other reported implementations~\cite{krishnamoorthi2018quantizing,liu2019learning}.
In addition, Table \ref{tab:int4_mobilenet} also reports the memory footprint of our PC+ICN INT4, which results to be 10\% less memory-demanding than using the integer thresholds based methodology.

%
%
%
%
\begin{figure}[]
\vskip 0.2in
\begin{center}
\centerline{\includegraphics[width=0.8\columnwidth]{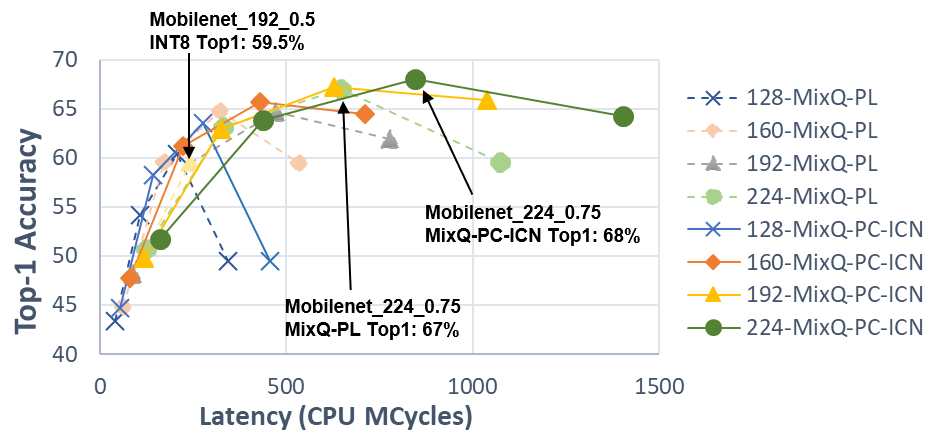}}
\caption{Accuracy-latency tradeoff of Mixed-Precision MobilenetV1 networks running on a STM32H7 device with $\mathrm{M_{RO}}=2MB$ and $\mathrm{M_{RW}}=512kB$.}
\label{fig:pareto}
\end{center}
\vskip -0.2in
\end{figure}

To evaluate our proposed methodology for the deployment of deep networks on microcontrollers, we apply our mixed-precision technique on all the Mobilenet configurations after setting the memory constraints $\mathrm{M_{RO}}=2MB$ and $\mathrm{M_{RW}}=512kB$, corresponding with the memory characteristics of an STM32H7 device.
The trained integer-only models are also bechmarked on the STM32H7 MCU running at 400MHz, to assess the implications for inference deployments. To this aim, we leverages an extended version of the ARM CMSIS-NN~\cite{lai2018cmsis} library, featuring an output stationary dataflow, and we measure latency in terms of clock cycles.
Figure~\ref{fig:pareto} plots the accuracy-latency tradeoff measured on two configurations. MixQ-PL indicates per-layer quantization with either the folding of batch-norm parameters or ICN for layers with $Q_y<8$ or $Q_w<8$. On the contrary, MixQ-PC-ICN indicates integer-only models with per-channel quantization and ICN as activation layers. Every curve represents a group of Mobilenet models with same input resolution. Increasing the width multiplier causes a longer latency because of the increasing amount of MAC operations.
When applying our mixed-precision method under this memory constraints, Mobilenet models with width multipliers of 0.25 and 0.5, with the exception of 224\_0.5, features no cuts of bit precision. Hence, under the configuration MixQ-PL, these points corresponds to the 8 bit integer-only models described in~\cite{jacob2018quantization}.
Pareto frontiers are mostly populated by MixQ-PC-ICN configurations. 
The most accurate model, PC+ICN 192\_0.5, scores 68\% Top1 accuracy by featuring 4 bit weight on the last convolutional pointwise and on the linear layers, in addition to $Q^1_y,Q^2_y,Q^5_y=4$, as determined by the memory-driven procedure of Section~\ref{sec:mem-driven}. 
This score is 8\% higher than the more accurate INT8 Mobilenet (192\_0.5) fitting into the same device.
Note that all the configurations featuring width multiplier $1.0$ suffers of a dramatic accuracy degradation with respect to full-precision settings (from 2\% to 15\%) due to aggressive quantization required to fit into the memory constrains.
On the latency side, the fastest inference model (128\_0.25 MixQ-PL), which features a homogeneous 8 bit quantization, runs at 10fps, 20$\times$ higher than the the most precise configuration (224\_0.75 PC+ICN), but only achieves 43\% of Top1 accuracy.
%
We can observe that the MixQ-PC-ICN quantization introduces a latency overhead of approx. 20\% with respect to the MixQ-PL setting, due to the additional subtractions of $Z_w$ biases within the inner loop of the convolution. On the other hand, MixQ-PC-ICN provides up to 4\% more accuracy for classification.
 
To further test our proposed mixed-precision method, we set the memory constrain to $\mathrm{M_{RO}}=1MB$ and compare with other mixed-precision methodologies in
Table \ref{tab:mixed_models}. 
Our best models feature up to 7\% lower accuracy with respect to~\cite{wang2018haq}, but we remark the integer-only nature of our solution. On the other hand, our implementation features a 2\% higher accuracy than INT8 models with comparable memory footprint and tailored for integer-only deployments. 

\begin{table}
\footnotesize
  \caption{Comparison with state-of-the-art mixed precision models when $M_{RO}$ is 1MB}
  \label{tab:mixed_models}
  \centering
  \resizebox{0.8\columnwidth}{!}{%
  \begin{tabular}{ l | l  | c  | c }
    \toprule
    Model & Quantization Method & Top1 Accuracy & Memory Constraints \\
    \midrule
    MobilenetV1\_224\_0.5 & MixQ-PC-ICN & 62.9\% & 1MB $M_{RO}$ + 512kB $M_{RW}$ \\
    MobilenetV1\_192\_0.5 & MixQ-PC-ICN & 60.2\% & 1MB $M_{RO}$ + 256kB $M_{RW}$ \\  
    \midrule
    MobilenetV1\_224\_0.5~\cite{jacob2018quantization}   & INT8 PL+FB       & 60.7\% & 1.34 MB \\ 
    MobilenetV1\_224\_0.25~\cite{jacob2018quantization}  & INT8 PL+FB       & 48.0\% & 0.47 MB \\ 
    \midrule
    MobilenetV1~\cite{wang2018haq} & MIX \textit{not-uniform}  & 57.14\% / 67.66\% & 1.09 / 1.58 MB \\ 
    MobileNetV2~\cite{wang2018haq} & MIX \textit{not-uniform} & 66.75\% / 70.90\% & 0.95 / 1.38 MB \\ 
    SqueezeNext~\cite{dong2019hawq} & MIX \textit{not-uniform} & 68.02\% & 1.09 MB \\ 
    \bottomrule
  \end{tabular}
  }
\end{table}

\section{Conclusion}
By mixing quantization methodologies is possible to execute complex deep neural networks such as MobilenetV1 on memory constrained MCU edge devices. To pursue this objective, in this work we introduced a mixed-precision quantization technique tailored for memory-constrained microcontroller devices, leveraging the formulation of a quantized activation layer, i.e. the Integer Channel-Normalization activation, to enable sub byte integer-only deployments.
The experimental results show a MobilenetV1 network running on a microcontroller equipped with 2MB of Flash and 512kB of RAM and featuring a Top1 accuracy of 68\%, which is 8\% higher  state-of-the-art integer-only 8 bit implementations.
%
%

\subsubsection*{Acknowledgments}
We thank the Italian Supercomputing Center CINECA for the access to their HPC facilities needed to run deep-learning experiments.

\medskip

\bibliographystyle{abbrvnat}

\bibliography{nips_refs}

\newpage

\appendix
\appendixpage

\section{Mixed-precision Quantization}
\begin{figure}[h]
\vskip 0.2in
\begin{center}
\centerline{\includegraphics[width=\columnwidth]{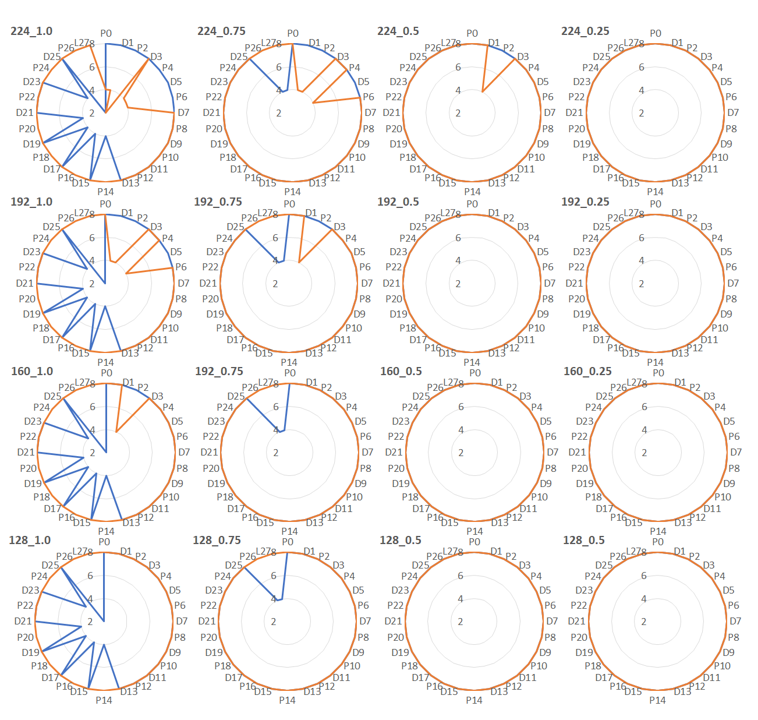}}
\caption{Weights (blue) and activations (orange) bit-precision of Mixed-Precision MobilenetV1 memory constraints with $\mathrm{M_{RO}}=2MB$ and $\mathrm{M_{RW}}=512kB$.}
\label{fig:bit_prec}
\end{center}
\vskip -0.2in
\end{figure}

Figure \ref{fig:bit_prec} plots the bit precision of every weight and activation tensor of the MixQ-PL and MixQ-PC-ICN MobilenetV1 models of experimental Section \ref{Sec:exp}. 
Table \ref{tab:mixed_models} reports the Top1 accuracy metrics of the experimented models.

\begin{table}
\footnotesize
  \caption{Top1 Accuracy of Mixed-Precision MobilenetV1 models}
  \label{tab:mixed_models}
  \centering
  \resizebox{0.8\columnwidth}{!}{%
  \begin{tabular}{ l | c  | c }
    \toprule
    Model & MixQ-PL Top1 Accuracy &  MixQ-PC-ICN Top1 Accuracy \\
    \midrule
    224\_1.0  & 59.61\% & 64.29\% \\
    224\_0.75 & \textbf{67.06\%} & \textbf{68.02\%} \\
    224\_0.5  & 63.12\% & 63.48\% \\
    224\_0.25 & 50.76\% & 51.70\% \\
    \midrule
    192\_1.0  & 61.94\% & 65.88\% \\
    192\_0.75 & 64.67\% & 67.23\% \\
    192\_0.5  & 59.50\% & 62.93\% \\
    192\_0.25 & 48.12\% & 49.75\% \\
    \midrule
    160\_1.0  & 59.49\% & 64.46\% \\
    160\_0.75 & 64.75\% & 65.70\% \\
    160\_0.5  & 59.55\% & 61.25\% \\
    160\_0.25 & 44.77\% & 47.79\% \\
    \midrule
    128\_1.0  & 49.44\% & 49.44\% \\
    128\_0.75 & 60.44\% & 63.53\% \\
    128\_0.5  & 54.20\% & 58.22\% \\
    128\_0.25 & 43.45\% & 44.68\% \\

    \bottomrule
  \end{tabular}
  }
\end{table}

\end{document}